\pdfoutput=1

\documentclass[11pt]{article}

\usepackage{emnlp2021}

\usepackage{times}
\usepackage{latexsym}
\usepackage{mathtools}
\usepackage[super]{nth}
\usepackage{amssymb}

\usepackage[T1]{fontenc}

\usepackage[utf8]{inputenc}

\usepackage{microtype}

\title{{A}ctive Learning for Argument Strength Estimation}

\author{Nataliia Kees \and Michael Fromm \and Evgeniy Faerman \and Thomas Seidl \\ LMU Munich, Germany \\ \texttt{kees.nataliia@gmail.com, fromm@dbs.ifi.lmu.de}}
\begin{document}
\maketitle
\begin{abstract}
High-quality arguments are an essential part of decision-making. Automatically predicting the quality of an argument is a complex task that recently got much attention in argument mining. However, the annotation effort for this task is exceptionally high. Therefore, we test uncertainty-based active learning (AL) methods on two popular argument-strength data sets to estimate whether sample-efficient learning can be enabled. Our extensive empirical evaluation shows that uncertainty-based acquisition functions can not surpass the accuracy reached with the random acquisition on these data sets.
\end{abstract}

\section{Introduction}
Argumentative quality plays a significant role in different domains of social activity where information and idea exchange are essential, such as the public domain and the scientific world. Theoretical discussions about what constitutes a good argument can be traced back to the ancient Greeks \cite{sep-aristotle-logic}. Researchers nowadays continue exploring this topic, trying out approaches that employ empirical machine learning estimation techniques \cite{simpson-gurevych-2018-finding}.

One of the most expensive and time-consuming tasks for machine learning-driven argument strength prediction is data labeling. Here, the result is highly dependent on the quality of labels, while the annotation task demands cognitive and reasoning abilities. One way to guarantee good annotations is to perform labeling with schooled experts, raising project costs extensively. For this reason, a common approach involves employing crowd workers. As argument strength detection is a highly subjective task, crowd workers' labeling results are often identified by low reliability and prompt researchers to counter-check the results with more crowd workers and as specifically developed agreement-based techniques. Sometimes a threshold for agreement cannot be reached at all, which might lead to data loss (see e.g. \cite{habernal-gurevych-2016-empirical, toledo2019automatic}.

This motivates us to investigate the applicability of some existing methods for reducing the amount of training data for automatic argument strength prediction. To this end, we look closely at the technique of active learning (AL). In this paper, we evaluate standard uncertainty-based acquisition functions for the argument strength prediction. We perform several experiments for the task of binary argument-pair classification (see Table \ref{example}) with several uncertainty-based data selection rounds. Our findings show that uncertainty-based AL techniques do not provide any advantages compared to random selection strategies. The cold-start problem and unreliable nature of annotations concerning argument strength might constitute the reasons for the failure of these techniques.

\begin{table}[h]
	\begin{tabular} {p{3.5cm} | p{3.5cm} } 
		\textbf{Argument 1} & \textbf{Argument 2}\\ 
		\hline
		School uniforms are a BAD idea. I'm to lazy to explain it but trust me, I wore them 4 years. & School uniform cant save person out of cold or heat like special clothes. It is not comfortable when you sit for an hours in a classroom.\\
	\end{tabular}
		\caption{Example of an argument pair both arguing against school uniforms \cite{habernal-gurevych-2016-argument}}
		\label{example}
\end{table}

\section{Related Work}
\subsection{Argument Quality Estimation}
In general, there is no agreement on how to operationalize argumentation quality \cite{toledo2019automatic, wachsmuth-etal-2017-computational, simpson-gurevych-2018-finding, persing2015modeling, lauscher2020rhetoric}. In some studies, argument strength is regarded in its persuasiveness and quantified as the proportion of people persuaded by the given argument \cite{habernal-gurevych-2016-argument, persing2015modeling, toledo2019automatic}. Persuasiveness makes argument strength easy to operationalize and serves as a way of dealing with the unclear nature of the concept by approximating its meaning through relying on the majority's wisdom. This approach lies at the center of the crowd-sourcing data labeling efforts and is the most common approach undertaken in existing data sets. This limits the reliability of the labels attained in such a manner, though, due to the highly subjective nature of such labels.

\subsection{Active Learning}

\emph{Active learning} is defined as a machine learning technique designed to assist in annotating unlabelled data sets by automatically selecting the most informative examples, which are subsequently labeled by human experts (the so-called \textit{oracles}) \cite{rongHu_text_class, cohn1996activelearning}. A popular approach to estimating the informativeness of single data points involves quantifying model uncertainty from a sample of stochastic forward passes for a given data point. Common techniques such as entropy, mutual information, or variation ratios (see Appendix \ref{appendix-acq-funcs} for more details) reportedly help reach good results on a range of tasks on high-dimensional data, e.g., in Computer Vision or Natural Language Processing \cite{gal2017imagedata, lipton2018active-nlp, rongHu_text_class}.
The assumption behind this is that in this way, data points which are closest to the decision boundary can be selected, helping to fine-tune the line dividing the classes most efficiently.

So far, AL in argument mining has received little attention. In the work of \cite{Ein-Dor_Shnarch_Dankin_Halfon_Sznajder_Gera_Alzate_Gleize_Choshen_Hou_Bilu_Aharonov_Slonim_2020}, the authors propose an Iterative Retrospective Learning (IRL) variant for the argument mining task. Their approach, however, is focused on solving the class imbalance problem between arguments and non-arguments and is \emph{precision-} rather than \emph{accuracy-oriented} as AL is. Another approach is suggested by \cite{simpson-gurevych-2018-finding}. They apply the Gaussian process preference learning (GPPL) method for performing AL for estimating argument convincingness, which the authors expect to be helpful against the cold-start problem. 

\section{Data Set}
\label{sec:dataset}
For our analysis, we use two publicly available data sets suitable for the task of pairwise argument strength prediction:
\begin{itemize}
	\item \textit{UKPConvArg1Strict}, published by \cite{habernal-gurevych-2016-argument}, consists of 11,650 argument pairs distributed over 16 topics. 
	\item \textit{IBM-9.1kPairs}, presented by \cite{toledo2019automatic} consists of 9,125 argument pairs distributed over 11 topics. 
\end{itemize} 

Because supporting and opposing arguments often share the same vocabulary and semantics, we do not treat each stance within a given topic as a separate topic, contrary to the authors of the two data sets. Instead, we combine the "for" and "against" arguments within the same topic under the same topic index and, thus, avoid leakage of semantic information between train and test data split. This preprocessing makes the performance of our models not directly comparable with the performance from the original papers. However, reproducibility of the original papers' results is beyond the scope of this work, as our focus lies on testing AL acquisition functions instead of reaching higher performance with our models.

Due to the high computational costs of the AL process, we decide to select the three most representative topics from each data set. One way to reach high representativeness would be to select topics that are average in difficulty. Since we try to approximate a real-world setting where the labels are unknown, it is not clear at the beginning which topics are more challenging to learn than the others. For this reason, we decide to select our test topics according to their size. Thus, we cross-validate our models on each data set's smallest topic, the largest one, and the median-sized one. Thus, the topics we select according to this procedure are topics 10 ("Is the school uniform a good or bad idea?"), 13 ("TV is better than books") and 14 ("Personal pursuit or advancing the common good?") in \textit{UKPConvArg1Strict} and topics 3 ("Does social media bring more harm than good?"), 4 ("Should we adopt cryptocurrency?") and 7 ("Should we ban fossil fuels?") in \textit{IBM-9.1kPairs} data.

\section{Experimental Setting}

\subsection{Research Design}
This study aims to test the hypothesis that \emph{uncertainty-based data acquisition strategies can help to achieve a better model performance than a mere random selection of the data for argument strength estimation}. We test this by comparing different data selection strategies against random data selection, serving as a baseline. 

We test our acquisition strategies on a task of pairwise (relative) argument strength comparison, constructed as a binary classification task, for which we use the \textit{UKPConvArg1Strict} \cite{habernal-gurevych-2016-argument} and \textit{IBM-9.1kPairs} \cite{toledo2019automatic} data sets. The code to our experiments is publicly available.\footnote{\url{https://github.com/nkees/active-learning-argument-strength}}

In order to employ uncertainty-based acquisition functions, we need to measure model uncertainty at prediction time. This is possible either by using Bayesian methods or by approximating their effect via obtaining distributions for output predictions by some other means. Based on the ground work layed out by \cite{gal2016dropout}, who show that dropout training in deep neural networks help approximate Bayesian inference in deep Gaussian processes, we design our experiments as MC dropout. With this, we simulate several stochastic forward passes through the model at prediction time and sample repeatedly from softmax outputs to obtain prediction distributions.

\subsection{Method and Procedure}

Similar to the procedure stipulated by \cite{toledo2019automatic}, we fine-tune the pre-trained BERT-Base Uncased English \cite{bert-paper} for the task of binary argument-pair classification by adding a single classification layer on top. The BERT architecture includes dropout layers with a probability of 0.1 \cite{bert-paper}. We keep it this way, which allows us to approximate model uncertainty as described above and test the uncertainty-based acquisition functions on the fine-tuned BERT-based. To do that, we enable dropout at inference time.

In order to estimate topic difficulty and validate our topic selection procedure described above, we train and test the models on all available labels of both data sets separately with the method of $k$-fold cross-validation, where $ k $ stands for the respective number of topics in a given data set. We separate every topic and use it as test data, with model training performed on the rest of the data, which helps to isolate the topics and measure their respective difficulty.

Our active learning experiments are conducted in a setting of a $3$-fold cross-validation, with 3 indicating the number of most representative topics selected by us from the given data sets, as mentioned in Section \ref{sec:dataset}. Thus, in each fold in our experiments, we test on one of the three selected topics for each data set (holdout data) and train on the rest of the compete data set (train-dev). 

The train-dev data in each fold consists of random splits into train (85\%) and validation (15\%) data, whereas the validation, or development, data are used for measuring the goodness of fit of the model trained on the training data. Having separated and fixed the validation data, a batch of 130 argument pairs is selected randomly from the train split. These data are used as initial training data on which bert-base-uncased is fine-tuned according to our classification task. 

Model evaluation is performed via accuracy measurement. Training on each of the three folds per data set is conducted ten times for improved reliability of the results. Thus, for each fold, we produce ten validation splits and ten initial training data batches to add some randomness into the experiments but in a controlled manner. They are kept fixed for every training fold to control for the effect of random initial data selection and enable a reliable comparison between the acquisition functions. 

We add another 130 argument pairs in each learning round and re-train the fine-tuned model. Within this setting, the whole data set would be selected within approx. 55 iterations for \textit{IBM-9.1kPairs} data and approx. 72 iterations for \textit{UKPConvArg1Strict} data (when calculated with the median-sized test split size). In an attempt to minimize the burden associated with heavy training, we decide to limit each active learning process to (less than) a half iterations, stopping at the \nth{27} iteration. 

Further details on the hyperparameters and the computing and software infrastructure can be found in Appendix sections \ref{app:infrastructure} and \ref{app:hyperparams}.

\subsection{Acquisition Functions}
We perform AL on three uncertainty-based acquisition functions one by one. In particular, we compare the performance of variation ratios, entropy, and BALD \cite{houlsby2011bayesian, gal2017imagedata} against a random acquisition baseline. For each of the learning rounds, we acquire data based on the heuristics calculated over a sample of 20 stochastic forward pass outputs. Our expectation is that other measures will outperform the random acquisition. 

\section{Results}

For the estimation of the performance of the models trained on the whole data with k-fold cross-validation, we reach a comparable performance of our BERT-based binary classification technique on both of the data sets (average accuracy on \textit{UKPConvArg1Strict}: 0.76, on \textit{IBM-9.1kPairs}: 0.77). This is a slightly worse performance than \cite{toledo2019automatic} achieves with the same architecture; the reason could be attributed to a different topic attribution strategy, as well as to some differences in the used hardware or hyperparameters, such as batch size or the number of epochs. 

We find that the topics selected by us from the \textit{UKPConvArg1Strict} stand rather on the low end of difficulty, with model accuracy tending towards the upper end of the scale when validated on these topics: all of them are higher than the mean performance of 0.76 (see Appendix \ref{sec:topic-hardness} for more details). However, from the distribution point of view, two of the topics, namely 10 and 13, yield median model performance, making them, in our opinion, suitable representatives of the whole data. 

As for the \textit{IBM-9.1kPairs} data set, our selected topics produce on average comparable performance with the model performance on the whole topic set (accuracy of 0.776 vs. 0.77 respectively). They also represent the most difficult topic, the easiest topic, and one closely neighboring the median topic (accuracy of 0.78 being slightly higher than the median performance of 0.77). In this case, the selected topics provide a better representation of the whole data set and grant strong validity when it comes to generalizing the results of our experiments.

The series of experiments we conducted in order to test whether our proposed heuristics for AL data acquisition provide us with any significant improvement surprisingly do not reveal any heuristic which would perform better than in the case of a random acquisition. This is true both for \textit{UKPConvArg1Strict} and \textit{IBM-9.1kPairs} data; a detailed overview is presented in Tables \ref{table:resultsUKPactive} and \ref{table:resultsIBMactive}. Statistical significance of the results has been tested with a Wilcoxon signed-rank test, which provides a non-parametric alternative to the paired T-test and is more suitable due to the non-Gaussian distribution of the differences in the results.

\begin{table}
	\centering
	\begin{tabular} {c c c c} 
		\hline
		\textbf{Heuristic} & \textbf{Mean} & \textbf{Variat.} & \textbf{Avg.Diff.} \\ 
		\hline\hline
		random (b.) & 0.747 & 0.0881 & -\\
		\hline
		entropy & 0.7388 & 0.0925 & -0.0082\\
		\hline
		variation ratios & 0.7368 & 0.0922 & -0.0103\\
		\hline
		bald & 0.7377 & 0.0928 & -0.0093\\
		\hline
	\end{tabular}

	\caption[Results of active learning experiments on \textit{UKPConvArg1Strict}]{Results of active learning experiments on \textit{UKPConvArg1Strict}. Abbreviations: \emph{b.} stands for \emph{baseline}, \emph{variat.} stands for \emph{variation}, \emph{avg.diff.} stands for \emph{average difference}. Negative average difference means that the challenger heuristic has not outperformed the baseline.}
	\label{table:resultsUKPactive}
\end{table}

All heuristics result in performance that is lower than that of the random baseline. All of our results are statistically significant with p-values $\leq 0.0001 $.

\begin{table}
	\centering
	\begin{tabular} {c c c c} 
		\hline
		\textbf{Heuristic} & \textbf{Mean} & \textbf{Variat.} & \textbf{Avg.Diff.} \\ 
		\hline\hline
		random (b.) & 0.7491 & 0.0855 & -\\
		\hline
		entropy & 0.7414 & 0.0878 & -0.0077\\
		\hline
		variation ratios & 0.7377 & 0.0923 & -0.0114\\
		\hline
		bald & 0.7412 & 0.0882 & -0.0079\\
		\hline
	\end{tabular}

\caption[Results of active learning experiments on \textit{IBM-9.1kPairs}]{Results of active learning experiments on \textit{IBM-9.1kPairs}. Abbreviations: \emph{b.} stands for \emph{baseline}, \emph{variat.} stands for \emph{variation}, \emph{avg.diff.} stands for \emph{average difference}. Negative average difference means that the challenger heuristic has not outperformed the baseline.}
\label{table:resultsIBMactive}
\end{table}

\begin{figure}[h]
\includegraphics[width=0.5\textwidth]{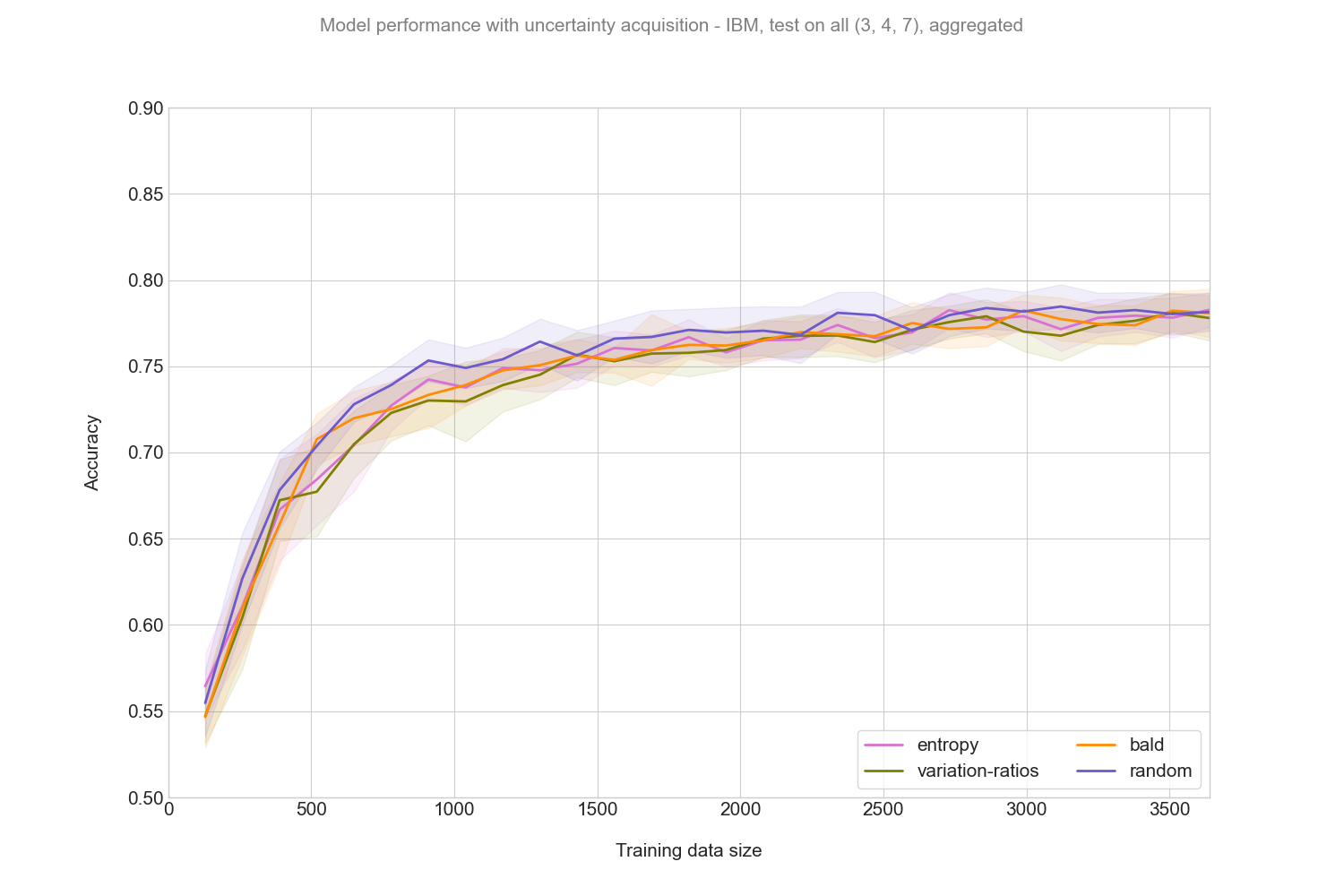}
\caption{Overview of the training results on the \textit{UKPConvArg1Strict} dataset based on different uncertainty-based acquisation methods}
\label{fig:uncerUKP}
\end{figure}

\begin{figure}[h]
\includegraphics[width=0.5\textwidth]{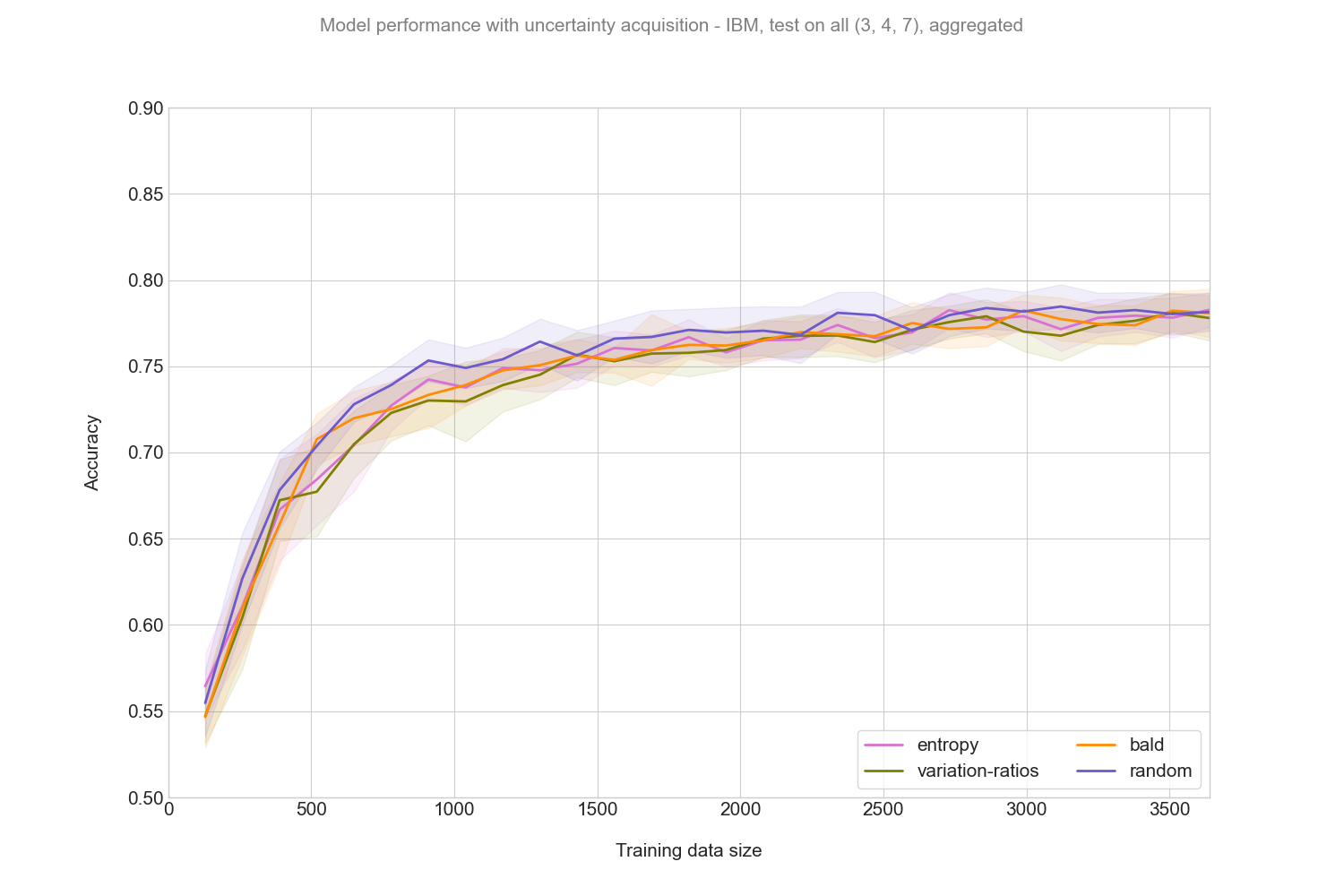}
\caption{Overview of the training results on the \textit{IBM-9.1kPairs} dataset based on different uncertainty-based acquisition methods}
\label{fig:uncerIBM}
\end{figure}

Despite the fact that random acquisition turns out to be the best one in terms of performance, with our results being consistent through both data sets and the difference being statistically significant, it is still noticeable that the differences in each case are rather small (see Figures \ref{fig:uncerUKP} and \ref{fig:uncerIBM} for graphic visualization of the model performance during active learning rounds comparing the acquisition functions).

\section{Discussion}
The results of our experiments do not point to any acquisition functions which outperform random acquisition. This finding does not exclude the possible existence of some other suitable acquisition functions, even from the same class (such as uncertainty-based). This remains an open question and should be considered in further research on the topic. For the time being, the random acquisition should be considered the approach of choice when selecting data for labeling for the task of pairwise argument strength prediction. This is sensible both from an accuracy standpoint as well as due to the computational cheapness of a random process. 

As the literature suggests, a possible reason why uncertainty-based methods perform so unimpressively is their proneness to picking outliers -- a disadvantage that some other methods, such as diversity-based acquisition (e.g., \cite{sener2018active}, do not have. This might be especially critical in the realm of argument strength prediction, as outliers might represent the arguments where relative argument strength difference is marginal, the data are noisy, or where the provided labeling is too subjective. Another critical factor is the cold-start problem, i.e., overfitting on the small initial data set of data, for which no initial informativeness estimation could be performed. This poses a drawback for the uncertainty-based methods, relying on the initial data sample for subsequent data acquisition.

\section{Conclusion}
This paper evaluates the effect of uncertainty-based acquisition functions, such as variation ratios, entropy, and BALD, on the model performance in the realm of argument strength prediction. As no acquisition function tested helps improve model performance in comparison to the random acquisition, we have not found any justification for using uncertainty-based active learning for pairwise argument strength estimation. 

\section{Acknowledgments}
This work has been funded by the German Federal Ministry of Education and Research (BMBF) under Grant No. 01IS18036A and by the Deutsche Forschungsgemeinschaft (DFG) within the project Relational Machine Learning for Argument Validation (ReMLAV), Grant Number SE 1039/10-1, as part of the Priority Program "Robust Argumentation Machines (RATIO)" (SPP-1999).
The authors of this work take full responsibility for its content.

\bibliography{dbstmpl}
\bibliographystyle{acl_natbib}

\appendix

\section{Appendix}
\label{sec:appendix}

\subsection{Uncertainty-based Acquisition Functions}
\label{appendix-acq-funcs}

In our work, we refer in particular to the following uncertainty-based acquisition functions \cite{gal2017imagedata}:

\begin{itemize}
    \item \textbf{variation ratios}: given a set of labels $ y_T $  from $ T $ stochastic forward passes, variation ratio for a given input point is calculated as:
        \begin{equation}
        varrat(x) = 1 - \frac{f_x}{T} 
        \end{equation}
    with $ f_x $ denoting the number of times the most commonly occurring category (mode of the distribution) has been sampled. This serves as an indication of how concentrated the predictions are, with 0.5 being the highest dispersion, i.e. uncertainty, and 0 being the highest concentration (certainty) in the case of binary classification.
    
    \item \textbf{predictive entropy}: stems from information theory and is calculated by averaging the softmax values for each class :

        \begin{equation}
        \begin{multlined}
        	predentr(x) = -\sum_{c}^{}p(y=c| \textbf{x}, D_{train})\\ \times \log_2(p(y=c| \textbf{x}, D_{train}))	,
        \end{multlined}
        \end{equation}
where $ p (y=c| \textbf{x}, D_{train}) $ stands for average probability of a data point adhering to a specific class given the outputs of the stochastic forward passes and the training data. $ c $ denotes the label class, i.e. we sum the values over all the classes to receive a  measure of entropy for a given data point.

\item \textbf{Bayesian Active Learning by Disagreement (BALD)} \cite{houlsby2011bayesian}, also called \textbf{mutual information} \cite{Gal2016Uncertainty}, is a function of predictive entropy as described above and averaged predictive entropies that have been calculated separately for each output:

\begin{equation}
\begin{multlined}
bald(x) = -[\sum_{c}^{}p(y=c| \textbf{x}, D_{train})\\ \times \log(p(y=c| \textbf{x}, D_{train}))] + \\ 
  \mathbb{E}_{p(\omega | D_{train})}[\sum_{c}^{}p(y=c| \textbf{w}, \omega)\\ \times \log(p(y=c| \textbf{x}, \omega))].
\end{multlined}
\end{equation}
\end{itemize}
\subsection{Computing \& Software Infrastructure}
\label{app:infrastructure}
The experiments were conducted on a Ubuntu 18.04 system with an AMD Ryzen Processor with 16 CPU-Cores, 126 GB memory, and a single NVIDIA RTX 2080 GPU with 11 GB memory. 
We further used Python 3.7, PyTorch 1.4 and the Huggingface-Transformer library (2.11.0).

\subsection{Topic Size and Difficulty}
\label{sec:topic-hardness}

\begin{table}[h]
	\centering
	\resizebox{\columnwidth}{!}{%

		\begin{tabular}{c l c c} 
		\hline
		No. & Topic & Size & Acc. \\ 
		\hline\hline
		0 & Ban Plastic Water Bottles? & 688 & 0.86 \\ 
		\hline
		1 & Christianity or Atheism & 588 & 0.81 \\
		\hline
		2 & Evolution vs. Creation & 782 & 0.78 \\
		\hline
		3 & Firefox vs. Internet Explorer & 748 & 0.81 \\
		\hline
		4 & Gay marriage - right or wrong? & 851 & 0.8 \\ 
		\hline
		5 & Should parents use spanking? & 706 & 0.76 \\ 
		\hline
		6 & If your spouse committed murder, would you turn them in? & 687 & 0.67 \\
		\hline
		7 & India has the potential to lead the world & 822 & 0.81 \\
		\hline
		8 & Is it better to have a lousy father or to be fatherless? & 616 & 0.64 \\
		\hline
		9 & Is porn wrong? & 571 & 0.79 \\
		\hline
		\textit{10} & \textit{Is the school uniform a good or bad idea?} & \textit{878} & \textit{0.78} \\ 
		\hline
		11 & Pro choice vs. Pro life & 845 & 0.61 \\
		\hline
		12 & Should physical edu. be mandatory? & 568 & 0.74 \\
		\hline
		\textit{13} &\textit{TV is better than books} & \textit{747} & \textit{0.79} \\
		\hline
		\textit{14} & \textit{Personal pursuit or common good?} & \textit{733} & \textit{0.84} \\
		\hline
		15 & Farquhar as the founder of Singapore & 820 & 0.7 \\ 
		\hline
		\hline
		& \textbf{Total Size/Average Acc.} & \textbf{11 650} & \textbf{0.76} \\ 
		\hline
	\end{tabular}}

	\caption[Topic sizes in \textit{UKPConvArg1Strict}]{Topic sizes in \textit{UKPConvArg1Strict}. Topics are provided with their corresponding numbers and size within the data set, as well as our model's performance at test time. The topics selected for testing the acquisition functions have been highlighted in italics.}
	\label{table:ukp}
\end{table}

\begin{table}[h]
	\centering 
	\resizebox{\columnwidth}{!}{%
		\begin{tabular}{c l c c} 
			\hline
			No. & Topic & Size & Acc. \\ 
			\hline\hline
			0 & Should flu vaccinations be mandatory? & 731 & 0.75 \\ 
			\hline
			1 & Should gambling be banned? & 503 & 0.8 \\
			\hline
			2 & Does online shopping bring more harm than good? & 278 & 0.79 \\
			\hline
			\textit{3} & \textit{Does social media bring more harm than good?} & \textit{2587} & \textit{0.78} \\
			\hline
			\textit{4} & \textit{Should we adopt cryptocurrency?} & \textit{719} & \textit{0.82} \\ 
			\hline
			5 & Should we adopt vegetarianism? & 1073 & 0.77 \\ 
			\hline
			6 & Should we sale violent video games to minors? & 484 & 0.74 \\
			\hline
			\textit{7} & \textit{Should we ban fossil fuels?} & \textit{263} & \textit{0.73} \\
			\hline
			8 & Should we legalize doping in sport? & 737 & 0.77 \\
			\hline
			9 & Should we limit autonomous cars? & 1217 & 0.79 \\
			\hline
			10 & Should we support information privacy laws? & 533 & 0.77 \\
			\hline
			\hline
			& \textbf{Total Size/Average Acc.} & \textbf{9 125} & \textbf{0.77} \\ 
			\hline
			
		\end{tabular}}

	\caption[Topic sizes in \textit{IBM-9.1kPairs}]{Topic sizes in \textit{IBM-9.1kPairs}. Topics are provided with their corresponding numbers and size within the data set, as well as our model's performance at test time. The topics selected for testing the acquisition functions have been highlighted in italics.}
	\label{table:ibm}
	
\end{table}

\subsection{Hyperparameters}
\label{app:hyperparams}
For the evaluation we initialized all methods for \textbf{ten} runs with different seeds and reported the \textbf{mean accuracy score}.
We used early stopping with a patience of three on a pre-selected validation set for regularization.
As loss function we used weighted binary-cross-entropy for the (relative) Argument Strength task.

We train our models on top of the pre-trained BERT-Base uncased with a dropout probability of 0.1. Learning rate is $ 2^{-5} $ (same as in \cite{toledo2019automatic}). The batch size per GPU is 64 and the model is validated after every half epoch.

\end{document}